%% file: acl_latex.tex
\pdfoutput=1

\documentclass[11pt]{article}

\usepackage[]{acl}
\usepackage{times}
\usepackage{latexsym}
\usepackage{graphicx}
\usepackage{float}
\usepackage{amsmath}
\usepackage{multirow}
\usepackage[linesnumbered,ruled,vlined]{algorithm2e}
\usepackage{hyperref}
\usepackage{caption}
\usepackage{subcaption}
\usepackage{url}
\usepackage{tabularx,multirow}
\usepackage{indentfirst}
\usepackage[T1]{fontenc}

\usepackage[utf8]{inputenc}

\usepackage{microtype}

\usepackage{marvosym}
\title{Exploring the Impact of Negative Samples of Contrastive Learning: \\ A Case Study of Sentence Embedding}

\author{
	Rui Cao\textsuperscript{1, *} \and 
	Yihao Wang\textsuperscript{1, *} \and 
	Yuxin Liang\textsuperscript{1} \\ {\bf
	Ling Gao\textsuperscript{1, $\dagger$} \and 
	Jie Zheng\textsuperscript{1} \and 
	Jie Ren \textsuperscript{2} \and 
	Zheng Wang \textsuperscript{3}} \\
	\textsuperscript{1}Northwest University, \textsuperscript{2}Shaanxi Normal University, \and \textsuperscript{3}University of Leeds \\
	\texttt{\{caorui, wangyihao, liangyuxin\}@stumail.nwu.edu.cn}, \\
	\texttt{\{gl, jzheng\}@nwu.edu.cn},	\texttt{renjie@snnu.edu.cn}, \texttt{z.wang5@leeds.ac.uk} 
}

\begin{document}
\maketitle
\renewcommand{\thefootnote}{\fnsymbol{footnote}}
\footnotetext[1]{ Authors contributed equally to this manuscript.}
\footnotetext[2]{ Corresponding author.}
\renewcommand{\thefootnote}{\arabic{footnote}}
\begin{abstract}

Contrastive learning is emerging as a powerful technique for extracting knowledge from unlabeled data. This technique requires a balanced mixture of two ingredients: positive (similar) and negative (dissimilar) samples. This is typically achieved by maintaining a queue of negative samples during training. Prior works in the area typically uses a fixed-length negative sample queue, but how the negative sample size affects the model performance remains unclear. The opaque impact of the number of negative samples on performance when employing contrastive learning aroused our in-depth exploration. This paper presents a momentum contrastive learning model with negative sample queue for sentence embedding, namely MoCoSE. We add the prediction layer to the online branch to make the model asymmetric and together with EMA update mechanism of the target branch to prevent the model from collapsing. We define a maximum traceable distance metric, through which we learn to what extent the text contrastive learning benefits from the historical information of negative samples. Our experiments find that the best results are obtained when the maximum traceable distance is at a certain range, demonstrating that there is an optimal range of historical information for a negative sample queue. We evaluate the proposed unsupervised MoCoSE on the semantic text similarity (STS) task and obtain an average Spearman's correlation of $77.27\%$. Source code is available \href{https://github.com/xbdxwyh/mocose}{here}.

\end{abstract}

\input{sections/intro}

\input{sections/related_work}
\input{sections/method}
\input{sections/exp}
\input{sections/empirical_study}
\input{sections/conclusion}
\section*{Acknowledgments}
Our work is supported by the National Key Research and Development Program of China under grant No.2019YFC1521400, National Natural Science Foundation of China under grant No.62072362 and No.61902229 and International Science and Technology Cooperation Project of Shaanxi (2020KW-006).

\bibliography{anthology,custom}
\bibliographystyle{acl_natbib}


\appendix
\input{sections/appendix}

\end{document}

%% file: sections/intro.tex
\section{Introduction}
%
In recent years, unsupervised learning has been brought to the fore in deep learning due to its ability to leverage large-scale unlabeled data. Various unsupervised contrastive models is emerging, continuously narrowing down the gap between supervised and unsupervised learning. Contrastive learning suffers from the problem of model collapse, where the model converges to a constant value and the samples all mapped to a single point in the feature space. Negative samples are an effective way to solve this problem.

In computer vision, SimCLR from Chen \cite{chen2020simple} and MoCo from He \cite{he2020momentum} is known for using negative samples and get the leading performance in the contrastive learning. SimCLR uses different data augmentation (e.g., rotation, masking, etc.) on the same image to construct positive samples, and negative samples are from the rest of images in the same batch. MoCo goes a step further by randomly select the data in entire unlabeled training set to stack up a first-in-first-out negative sample queue.

Recently in natural language processing, contrastive learning has been widely used in the task of learning sentence embedding. One of current state-of-the-art unsupervised method is SimCSE \cite{gao2021simcse}. Its core idea is to make similar sentences in the embedding space closer while keeping dissimilar away from each other. SimCSE uses dropout mask as augmentation to construct positive text sample pairs, and negative samples are picked from the rest of sentences in the same batch. The mask adopted from the standard Transformer makes good use of the minimal form of data augmentation brought by the dropout. Dropout results in a minimal difference without changing the semantics, reducing the negative noise introduced by augmentation.
However, the negative samples in SimCSE are selected from the same training batch with a limited batch size. Our further experiments show that SimCSE does not obtain improvement as the batch size increases, which arouses our interest in using the negative sample queue.

To better digging in the performance of contrastive learning on textual tasks, we build a contrastive model consisting of a two-branch structure and a negative sample queue, namely MoCoSE (\textbf{Mo}mentum \textbf{Co}ntrastive \textbf{S}entence \textbf{E}mbedding with negative sample queue). We also introduce the idea of asymmetric structure from BYOL \cite{grill2020bootstrap} by adding a prediction layer to the upper branch (i.e., the online branch). The lower branch (i.e., the target branch) is updated with exponential moving average (EMA) method during training. We set a negative sample queue and update it using the output of target branch.
Unlike directly using negative queue as in MoCo, for research purpose, we set an initialization process with a much smaller negative queue, and then filling the entire queue through training process, and update normally. We test both character-level (e.g., typo, back translation, paraphrase) and vector-level (e.g., dropout, shuffle, etc.) data augmentations and found that for text contrastive learning, the best results are obtained by using FGSM and dropout as augmentations.

Using the proposed MoCoSE model, we design a series of experiments to explore the contrastive learning for sentence embedding. We found that using different parts of samples from the negative queue leads to different performance. In order to test how much text contrastive learning benefit from historical information of the model, we proposed a maximum traceable distance metric. The metric calculates how many update steps before the negative samples in the queue are pushed in, and thus measures the historical information contained in the negative sample queue. We find that the best results can be achieved when the maximum traceable distance is within a certain range, reflected in the performance of uniformity and alignment of the learned text embedding. Which means there is an optimal interval for the length of negative sample queue in text contrastive learning model.

Our main contributions are as follows:

1. We combine several advantages of frameworks from image contrastive learning to build a more generic text unsupervised contrastive model. We carried out a detailed study of this model to achieve better results on textual data.

2. We evaluate the role of negative queue length and the historical information that the queue contains in text contrastive learning. By slicing the negative sample queue and using different positions of negative samples, we found those near the middle of the queue provides a better performance.

3. We define a metric called 'maximum traceable distance' to help analyze the impact of negative sample queue by combining the queue length, EMA parameter, and batch size. We found that changes in MTD reflects in the performance of uniformity and alignment of the learned text embedding.

%% file: sections/related_work.tex
\section{Related Work}
\textbf{Contrastive Learning in CV}

Contrast learning is a trending and effective unsupervised learning framework that was first applied to the computer vision \cite{hadsell2006dimensionality}. The core idea is to make the features of images within the same category closer and the features in different categories farther apart. 
Most of the current work are using two-branch structure \cite{chen2021jigsaw}. While influential works like SimCLR and MoCo using positive and negative sample pairs, BYOL \cite{grill2020bootstrap} and SimSiam \cite{chen2021exploring} can achieve the same great results with only positive samples. BYOL finds that by adding a prediction layer to the online branch to form an asymmetric structure and using momentum moving average to update the target branch, can train the model using only positive samples and avoid model collapsing. SimSiam explores the possibility of asymmetric structures likewise. Therefore, our work introduces this asymmetric idea to the text contrastive learning to prevent model collapse. In addition to the asymmetric structure and the EMA mechanism to avoid model collapse, some works consider merging the constraint into the loss function, like Barlow Twins \cite{zbontar2021barlow}, W-MSE \cite{ermolov2021whitening}, and ProtoNCE \cite{li2021prototypical}.

\textbf{Contrastive Learning in NLP}

Since BERT \cite{devlin2018bert} redefined state-of-the-art in NLP, leveraging the BERT model to obtain better sentence representation has become a common task in NLP. A straightforward way to get sentence embedding is by the $[CLS]$ token due to the Next Sentence Prediction task of BERT. But the $[CLS]$ embedding is non-smooth anisotropic in semantic space, which is not conducive to STS tasks, this is known as the representation degradation problem \cite{gao2019representation}. BERT-Flow \cite{li2020on} and BERT-whitening \cite{su2021whitening} solve the degradation problem by post-processing the output of BERT. SimCSE found that utilizing contrasting mechanism can also alleviate this problem.

Data augmentation is crucial for contrastive learning. 
In CLEAR \cite{wu2020clear}, word and phrase deletion, phrase order switching, synonym substitution is served as augmentation. CERT \cite{fang2020cert} mainly using back-and-forth translation, and CLINE \cite{wang2021cline} proposed synonym substitution as positive samples and antonym substitution as negative samples, and then minimize the triplet loss between positive, negative cases as well as the original text. ConSERT \cite{yan2021consert} uses adversarial attack, token shuffling, cutoff, and dropout as data augmentation. CLAE \cite{ho2020contrastive} also introduces Fast Gradient Sign Method, an adversarial attack method, as text data augmentation. Several of these augmentations are also introduced in our work. The purpose of data augmentation is to create enough distinguishable positive and negative samples to allow contrastive loss to learn the nature of same data after different changes. Works like \cite{mitrovic2020less} points out that longer negative sample queues do not always give the best performance. This also interests us how the negative queue length affects the text contrastive learning.

%% file: sections/method.tex
\section{Method}
\begin{figure*}[htb]
	\centering
	\includegraphics[width=0.75\textwidth]{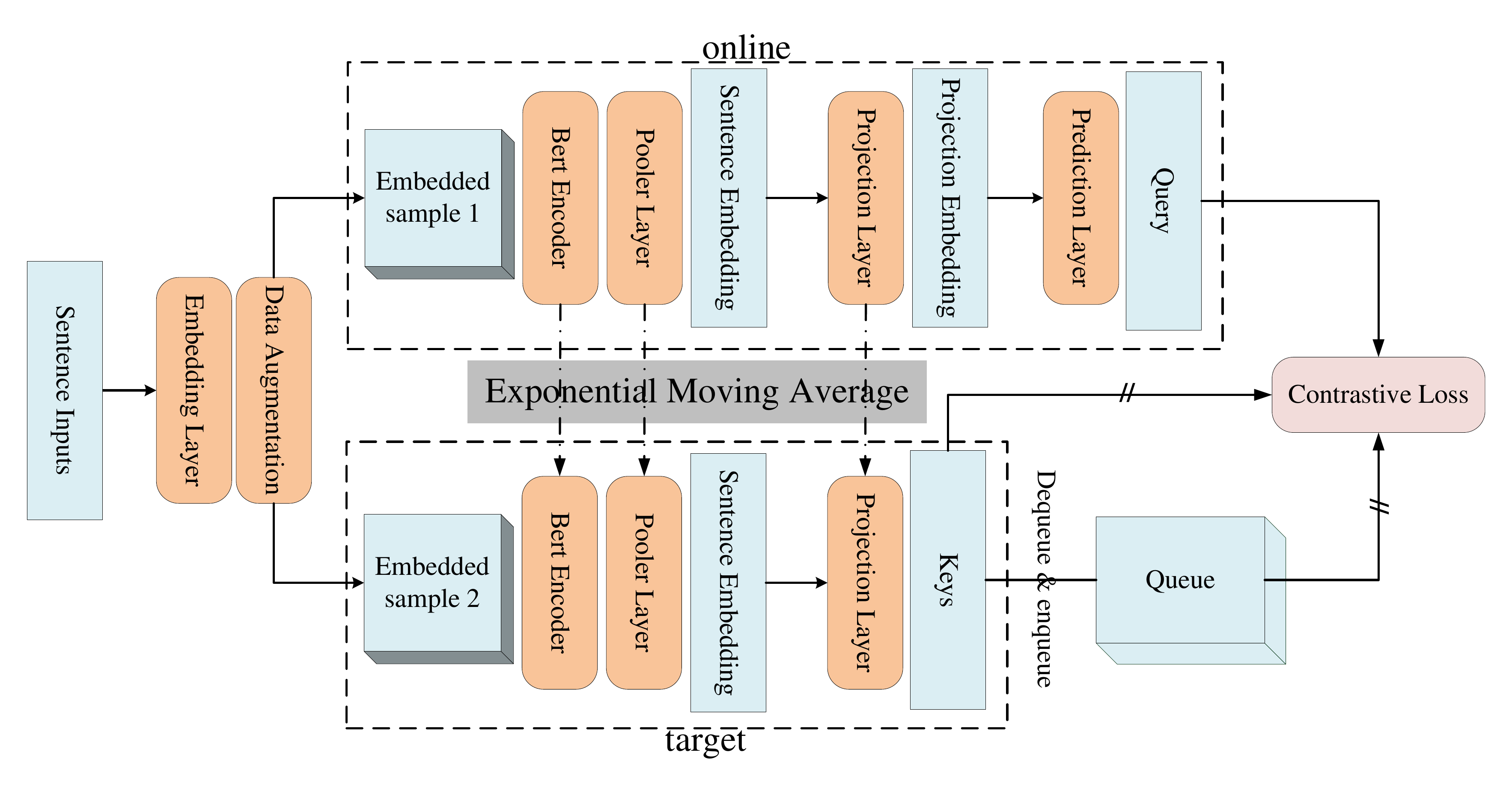}
	\caption{\label{fig1}The model structure of MoCoSE. The embedding layer consists of a BERT embedding layer with additional data augmentation. The pooler, projection, and predictor layers all keep the same dimensions with the encoder layer. The MoCoSE minimizes contrastive loss between query, queue and keys (i.e. InfoNCE loss).}
\end{figure*}

Figure \ref{fig1} depicts the architecture of proposed MoCoSE. In the embedding layer, two versions of the sentence embedding are generated through data augmentation ($ dropout=0.1$ and $fgsm=5e-9$). The resulting two slightly different embeddings then go through the online and target branch to obtain the query and key vectors respectively. The structure of encoder, pooler and projection of online and target branch is identical. We add a prediction layer to the online branch to make asymmetry between online and target branch. The pooler, projection and prediction layers are all composed of several fully connected layers.

Finally, the model calculates contrasting loss between query, key and negative queue to update the online branch. In the process, key vector serves as positive sample with respect to the query vector, while the sample from queue serves as negative sample to the query. The target branch truncates the gradient and updated with the EMA mechanism. The queue is a first-in-first-out collection of negative samples with size $K$ which means it sequentially stores the key vectors generated from the last few training steps.

The PyTorch style pseudo-code for training MoCoSE with the negative sample queue is shown in Algorithm \ref{alg:mocose} in Appendix \ref{A.2}.

\textbf{Data Augmentation}
Comparing with SimCSE, we tried popular methods in NLP such as paraphrasing, back translation, adding typos etc., but experiments show that only adversarial attacks and dropout have improved the results. We use FGSM \cite{goodfellow2015explaining} (Fast Gradient Sign Method) as adversarial attack. In a white-box environment, FGSM first calculates the derivative of model with respect to the input, and use a sign function to obtain its specific gradient direction. Then, after multiplying it by a step size, the resulting 'perturbation' is added to the original input to obtain the sample under the FGSM attack.
\begin{equation}
x'=x+\varepsilon \cdot sign\left ( \nabla_{x}\mathcal{L}\left ( x,\theta \right ) \right ) 
\end{equation}
Where $x$ is the input to the embedding layer, $\theta$ is the online branch of the model, and $\mathcal{L}(\cdot)$ is the contrastive loss computed by the query, key and negative sample queue. $\nabla_{x}$ is the gradient computed through the network for input $x$, $sign()$ is the sign function, and $\varepsilon$ is the perturbation parameter which it controls how much noise it added.

\textbf{EMA and Asymmetric Branches}
Our model uses EMA mechanism to update the target branch. Formally, denoting the parameters of online and target branch as $\theta_o$ and $\theta_t$, EMA decay weight as $\eta$, we update $\theta_t$ by:
\begin{equation}
\theta_t \leftarrow \eta \theta_t + (1-\eta) \theta_o
\end{equation}
Experiments demonstrate that not using EMA leads to model collapsing, which means the model did not converge during training. The prediction layer we added on the online branch makes two branches asymmetric to further prevent the model from collapsing. For more experiment details about symmetric model structure without EMA mechanism, please refer to Appendix \ref{A.1}.

\textbf{Negative Sample Queue}
The negative sample queue has been theoretically proven to be an effective means of preventing model from collapsing. Specifically, both the queue and the prediction layer of the upper branch serves to disperse the output feature of the upper and lower branches, thus ensuring that the contrastive loss obtains features with sufficient uniformity. We also set a buffer for the initialization of the queue, i.e., only a small portion of the queue is randomly initialized at the beginning, and then enqueue and dequeue normally until the end.


\textbf{Contrastive Loss}
Similar to MoCo, we also use InfoNCE \cite{oord2018representation} as contrastive loss, as shown in eq.(\ref{infonce}). 
\begin{equation}
	\begin{aligned}
		\mathcal{L} =-\log{\frac{exp\left ( q\cdot k/\tau \right ) }{exp\left ( q\cdot k/\tau \right )+ {\textstyle \sum_{l}exp\left ( q\cdot l/\tau  \right ) } } }
		\label{infonce}
	\end{aligned}
\end{equation} 

Where, $q$ refers to the query vectors obtained by the online branch; $k$ refers to the key vectors obtained by the target branch; and $l$ is the negative samples in the queue; $\tau$ is the temperature parameter.

%% file: sections/exp.tex
\section{Experiments}
\subsection{Settings}
We train with a randomly selected corpus of 1 million sentences from the English Wikipedia, and we conduct experiments on seven standard semantic text similarity (STS) tasks, including STS 2012—2016 \cite{agirre2012semeval,agirre2013sem,agirre2014semeval,agirre2015semeval,agirre2016semeval}, STSBenchmark \cite{cer2017semeval} and SICK-Relatedness \cite{wijnholds2021sick}. 
The SentEval\footnote{https://github.com/facebookresearch/SentEval} toolbox is used to evaluate our model, and we use the Spearman’s correlation to measure the performance. We start our training by loading pre-trained BERT checkpoints\footnote{https://huggingface.co/models} and use the $[CLS]$ token embedding from the model output as the sentence embedding. In addition to the semantic similarity task, we also evaluate on seven transfer learning tasks to test the generalization performance of the model. For text augmentation, we tried several vector-level methods mentioned in ConSERT, including position shuffle, token dropout, feature dropout. In addition, we also tried several text-level methods from the nlpaug\footnote{https://github.com/makcedward/nlpaug} toolkit, including synonym replace, typo, back translation and paraphrase.

\textbf{Training Details} The learning rate of MoCoSE-BERT-base is set to 3e-5, and for MoCoSE-BERT-large is 1e-5. With a weight decay of 1e-6, the batch size of the base model is 64, and the batch size of the large model is 32. We validate the model every 100 step and train for one epoch. The EMA decay weight $\eta$ is incremented from $0.75$ to $0.95$ by the cosine function. The negative queue size is 512. For more information please refer to Appendix \ref{A.0}.

\subsection{Main Results}

We compare the proposed MoCoSE with several commonly used unsupervised methods and the current state-of-the-art contrastive learning method on the text semantic similarity (STS) task, including average GloVe embeddings \cite{pennington-etal-2014-glove}, average BERT or RoBERTa embeddings, BERT-flow, BERT-whitening, ISBERT \cite{zhang2020unsupervised}, DeCLUTR \cite{giorgi-etal-2021-declutr}, CT-BERT \cite{carlsson2021semantic} and SimCSE.

\begin{table*}[htb]
	\centering
	\scalebox{0.85}{
	\begin{tabular}{lllllllll}
		\hline
		Model                           & STS12 & STS13 & STS14 & STS15 & STS16 & STS-B & SICK-R & Avg.   \\
		\hline
		\multicolumn{9}{c}{Unsupervised Models (Base)}                                                           \\
		\hline
		GloVe (avg.)                & 55.14 & 70.66 & 59.73 & 68.25 & 63.66 & 58.02 & 53.76  & 61.32  \\
		BERT (first-last avg.)      & 39.70 & 59.38 & 49.67 & 66.03 & 66.19 & 53.87 & 62.06  & 56.70  \\
		BERT-flow                   & 58.40 & 67.10 & 60.85 & 75.16 & 71.22 & 68.66 & 64.47  & 66.55  \\
		BERT-whitening              & 57.83 & 66.90 & 60.90 & 75.08 & 71.31 & 68.24 & 63.73  & 66.28  \\
		IS-BERT                     & 56.77 & 69.24 & 61.21 & 75.23 & 70.16 & 69.21 & 64.25  & 66.58  \\
		CT-BERT                     & 61.63 & 76.80 & 68.47 & 77.50 & 76.48 & 74.31 & 69.19  & 72.05  \\
		RoBERTa (first-last avg.)   & 40.88 & 58.74 & 49.07 & 65.63 & 61.48 & 58.55 & 61.63  & 56.57  \\
		RoBERTa-whitening           & 46.99 & 63.24 & 57.23 & 71.36 & 68.99 & 61.36 & 62.91  & 61.73  \\
		DeCLUTR-RoBERT              & 52.41 & 75.19 & 65.52 & 77.12 & 78.63 & 72.41 & 68.62  & 69.99  \\
		SimCSE                      & 68.40 & \textbf{82.41} & 74.38 & 80.91 & 78.56 & 76.85 & 72.23  & 76.25  \\
		MoCoSE                      & \textbf{71.48} & 81.40 & \textbf{74.47} & \textbf{83.45} & \textbf{78.99} & \textbf{78.68} & \textbf{72.44}  & \textbf{77.27}  \\
		\hline
		\multicolumn{9}{c}{Unsupervised Models (Large)}                                                   \\
		\hline
		SimCSE-RoBERTa             & 72.86 & 83.99 & 75.62 & \textbf{84.77} & \textbf{81.80} & \textbf{81.98} & 71.26  & 78.90  \\
		SimCSE-BERT                & 70.88 & 84.16 & 76.43 & 84.50 & 79.76 & 79.26 & \textbf{73.88}  & 78.41  \\
		MoCoSE-BERT                & \textbf{74.50} & \textbf{84.54} & \textbf{77.32} & 84.11 & 79.67 & 80.53 & 73.26  & \textbf{79.13}  \\
		\hline
	\end{tabular}
	}
	\caption{\label{table1}Spearman correlation of MoCoSE on seven semantic text similarity tasks. We compared with the state-of-the-art method SimCSE. MoCoSE achieves the best results with both BERT-base and BERT-large pre-trained models.}
\end{table*}
\begin{table*}[htb]
	\centering
	\scalebox{0.85}{
	\begin{tabular}{lllllllll}
		\hline
		Model                  & MR    & CR    & SUBJ  & MPQA  & SST   & TREC  & MRPC  & Avg.   \\
		\hline
		\multicolumn{9}{c}{Unsupervised Model (Base)}                                           \\
		\hline
		GloVe (avg.)           & 77.25 & 78.30 & 91.17 & 87.85 & 80.18 & 83.00  & 72.87 & 81.52  \\
		Skip-thought           & 76.50 & 80.10 & 93.60 & 87.10 & 82.00 & 92.20  & 73.00 & 83.50  \\
		Avg. BERT
		embeddings & 78.66 & 86.25 & 94.37 & 88.66 & 84.40 & \textbf{92.80} & 69.54 & 84.94  \\
		BERT-[CLS]embedding    & 78.68 & 84.85 & 94.21 & 88.23 & 84.13 & 91.40 & 71.13 & 84.66  \\
		SimCSE-RoBERTa         & 81.04 & \textbf{87.74} & 93.28 & 86.94 & \textbf{86.60} & 84.60 & 73.68 & 84.84  \\
		SimCSE-BERT            & \textbf{81.18} & 86.46 & 94.45 & 88.88 & 85.50 & 89.80 & 74.43 & \textbf{85.81}  \\
		MoCoSE-BERT                 & 81.07 & 86.43 & \textbf{94.76} & \textbf{89.70} & 86.35 & 84.06 & \textbf{75.86} & 85.46  \\
		\hline
		\multicolumn{9}{c}{Unsupervised Model (Large)}                                          \\
		\hline
		SimCSE-RoBERTa         & 82.74 & 87.87 & 93.66 & 88.22 & \textbf{88.58} & \textbf{92.00} & 69.68 & 86.11  \\
		MoCoSE-BERT            & \textbf{83.71} & \textbf{89.07} & \textbf{95.58} & \textbf{90.26} & 87.96 & 84.92 & \textbf{76.81} & \textbf{86.90}  \\
		\hline
	\end{tabular}
	}
	\caption{\label{table2}Performance of MoCoSE on the seven transfer tasks. We compare the performance of MoCoSE and other models on the seven transfer tasks evaluated by SentEval, and MoCoSE remains at a comparable level with the SimCSE.}
\end{table*}
As shown in Table \ref{table1}, the average Spearman's correlation of our best model is $77.27\%$, outperforming unsupervised SimCSE with BERT-base. Our model outperforms SimCSE on STS2012, STS2015, and STS-B, and SimCSE perform better on the STS2013 task. Our MoCoSE-BERT-large model outperforms SimCSE-BERT-Large by about $0.7$ on average, mainly on STS12, STS13, and STS14 tasks, and maintains a similar level on other tasks.

Furthermore, we also evaluate the performance of MoCoSE on the seven transfer tasks provided by SentEval. As shown in Table \ref{table2}, MoCoSE-BERT-base outperforms most of the previous unsupervised method, and is on par with SimCSE-BERT-base.

%% file: sections/empirical_study.tex
\section{Empirical Study}
To further explore the performance of the MoCo-like contrasting model on learning sentence embedding, we set up the following ablation trials.

\subsection{EMA Decay Weight}
We use EMA to update the model parameters for the target branch and find that EMA decay weight affects the performance of the model. The EMA decay weight affects the update process of the model, which further affects the vectors involved in the contrastive learning process. Therefore, we set different values of EMA decay weight and train the model with other hyperparameters held constant. As shown in Table \ref{table3} and Appendix \ref{A.4}, the best result is obtained when the decay weight of EMA is set to 0.85.
\begin{table}[htb]
	\centering
	\scalebox{0.85}{
	\begin{tabular}{lllllll} 
		\hline
		EMA    & 0.5       & 0.8      & \textbf{0.85}      & 0.9      & 0.95      & 0.99       \\ 
		\hline
		Avg.   & 75.76     & 75.19    & \textbf{76.49}     & 76.05    & 76.08     & 75.12      \\
		\hline
	\end{tabular}
	}
	\caption{\label{table3}Effect of EMA decay weight on model performance. The best results are obtained with the EMA decay weight at 0.85.}
\end{table}
Compared to the choice of EMA decay weight in CV (generally as large as $0.99$), the value of $0.85$ in our model is smaller, which means that the model is updated faster. We speculate that this is because the NLP model is more sensitive in the fine-tuning phase and the model weights change more after each step of the gradient, so a faster update speed is needed.

\subsection{Projection and Prediction}
Several papers have shown (e.g. Section F.1 in BYOL \cite{grill2020bootstrap}) that the structure of projection and prediction layers in a contrastive learning framework affects the performance of the model. We combine the structure of projection and prediction with different configurations and train them with the same hyperparameters. As shown in Table \ref{table4},  the best results are obtained when the projection is $1$ layer and the prediction has $2$ layers. The experiments also show that the removal of projection layers degrades the performance of the model.

\begin{table}[tb]
	\centering
	\scalebox{0.85}{
	\begin{tabular}{lllllll} 
		\cline{1-3}\cline{5-7}
		Proj. & Pred. & Corr.    &  & Proj. & Pred. & Corr.  \\ 
		\cline{1-3}\cline{5-7}
		& 1          & 60.46          &  &            & 1          & 66.96        \\
		0          & 2          & 62.67          &  & 2          & 2          & 66.29        \\
		& 3          & 63.62          &  &            & 3          & 61.57        \\ 
		\cline{1-3}\cline{5-7}
		& 1          & 76.74          &  &            & 1          & 31.51        \\
		\textbf{1} & \textbf{2} & \textbf{76.89} &  & 3          & 2          & 43.97        \\
		& 3          & 76.24          &  &            & 3          & 39.13        \\
		\cline{1-3}\cline{5-7}
	\end{tabular}
	}
	\caption{\label{table4}The impact of different combinations of projection and predictor on the model.}
\end{table}


\subsection{Data Augmentation}
We investigate the effect of some widely-used data augmentation methods on the model performance. As shown in Table \ref{table5}, cut off and token shuffle do not improve, even slightly hurt the model's performance. Only the adversarial attack (FGSM) has slight improvement on the performance. Therefore, in our experiments, we added FGSM as a default data augmentation of our model in addition to dropout. Please refer to Appendix \ref{A.6} for more FGSM parameters results.
\begin{table}[htb]
	\centering
	\scalebox{0.85}{
	\begin{tabular}{ll} 
		\hline
		Augmentation Methods        & Avg.   \\ 
		\hline
		Dropout only                & 76.76  \\
		+ \textbf{FGSM} ($\varepsilon$=5e-9)                      & \textbf{77.04}  \\
		+ Position\_shuffle (True)   & 73.80  \\
		+ Token   dropout (prob=0.1)    & 41.32  \\
		+ Feature dropout (prob=0.01) & 76.33  \\
		+ Feature dropout (prob=0.1)           & 71.62  \\
		+ Typos   & 22.32  \\
		+ Synonym replace (roberta-base)            & 28.70  \\
		+ Paraphrasing (xlnet-base-cased)           & 60.45  \\
		+ Backtranslation (en->de->en)           & 69.35  \\
		\hline
	\end{tabular}
	}
	\caption{\label{table5}The effect of different data augmentation methods.}
\end{table}
We speculate that the reason token cut off is detrimental to the model results is that the cut off perturbs too much the vector formed by the sentences passing through the embedding layer. Removing one word from the text may have a significant impact on the semantics. We tried two parameters 0.1 and 0.01 for the feature cut off, and with these two parameters, the results of using the feature cut off is at most the same as without using feature the cut off, so we discard the feature cut off method. More results can be found in Appendix \ref{A.5}.

\indent The token shuffle is slightly, but not significantly, detrimental to the results of the model. This may be due to that BERT is not sensitive to the position of token. In our experiment, the sentence-level augmentation methods also failed to outperform than the drop out, FGSM and position shuffle.

\indent Among the data augmentation methods, only FGSM together with dropout improves the results, which may due to the adversarial attack slightly enhances the difference between the two samples and therefore enables the model to learn a better representation in more difficult contrastive samples.

\subsection{Predictor Mapping Dimension}
The predictor maps the representation to a feature space of a certain dimension. We investigate the effect of the predictor mapping dimension on the model performance. Table \ref{table6}.a shows that the predictor mapping dimension can seriously impair the performance of the model when it is small, and when the dimension rises to a suitable range or larger, it no longer has a significant impact on the model. This may be related to the intrinsic dimension of the representation, which leads to the loss of semantic information in the representation when the predictor dimension is smaller than the intrinsic dimension of the feature, compromising the model performance. We keep the dimension of the predictor consistent with the encoder in our experiments. More results can be found in Appendix \ref{A.7}.
\begin{table}[htb]
	\centering
	\scalebox{0.85}{
	\begin{subtable}{0.42\linewidth}
		\centering
		\begin{tabular}{ll} 
			\hline
			Dim  & Avg.   \\ 
			\hline
			256  & 73.91  \\
			512  & 76.07  \\
			\textbf{768}  & \textbf{77.04}  \\
			1024 & 77.02  \\
			2048 & 77.03  \\
			\hline
		\end{tabular}
		\caption{\label{a}}
	\end{subtable}
	\begin{subtable}{0.42\linewidth}
		\centering
		\begin{tabular}{ll} 
			\hline
			Size & Avg.   \\ 
			\hline
			32   & 73.86  \\
			\textbf{64}   & \textbf{77.25}  \\
			128  & 76.78  \\
			256  & 76.62  \\
			\hline
		\end{tabular}
		\caption{\label{b}}
	\end{subtable}
	}
	\caption{\label{table6}(a) Impact of prediction dimension on model performance. (b) Impact of batch size on the model with fixed queue size. Both table under a batch size setting to 512.}
\end{table}

\subsection{Batch Size}
With a fixed queue size, we investigated the effect of batch size on model performance, the results is in Table \ref{table6}.b, and the model achieves the best performance when the batch size is 64. Surprisingly the model performance does not improve with increasing batch size, which contradicts the general experience in image contrastive learning. This is one of our motivations for further exploring the effect of the number of negative samples on the model. 

\subsection{Size of Negative Sample Queue}
The queue length determines the number of negative samples, which direct influence performance of the model. We first test the size of negative sample queue to the model performance. With queue size longer than 1024, the results get unstable and worse. We suppose this may be due to the random interference introduced to the training by filling the initial negative sample queue. This interference causes a degradation of the model's performance when the initial negative sample queue becomes longer. To reduce the drawbacks carried out by this randomness, we changed the way the negative queue is initialized. We initialize a smaller negative queue, then fill the queue to its set length in the first few updates, and then update normally. According to experiments, the model achieves the highest results when the negative queue size set to $512$ and the smaller initial queue size set to $128$.

According to the experiments of MoCo, the increase of queue length improves the model performance. However, as shown in Table \ref{table7}, increasing the queue length with a fixed batch size decreases our model performance, which is not consistent with the observation in MoCo. We speculate that this may be due to that NLP models updating faster, and thus larger queue lengths store too much outdated feature information, which is detrimental to the performance of the model. Combined with the observed effect of batch size, we further conjecture that the effect of the negative sample queue on model performance is controlled by the model history information contained in the negative sample in the queue. See Appendix \ref{A.8} and \ref{A.9} for more results of the effect of randomization size and queue length.

\begin{table}[tb]
	\centering
	\scalebox{0.8}{
	\begin{tabular}{llllll}
		\hline
		\multirow{2}{*}{\begin{tabular}[c]{@{}l@{}}Initial \\Size\end{tabular}}&
		\multicolumn{5}{c}{Queue Size} \\
		& 128   & 256   & 512   & 1024  & 4096 \\
		\hline
		w.o. init.           & 76.40          & 76.19          & 75.38          & \textbf{76.63} & 50.17          \\
		init. 1/4 queue       & 75.92          & 76.34          & \textbf{77.30} & 76.20          & \textbf{50.42} \\
		init. 1/2 queue       & 76.16          & \textbf{76.39} & 76.94          & 76.57          & 38.74          \\
		init. all (normal) & \textbf{76.87} & 75.81          & 76.29          & 76.45          & 45.80  \\
		\hline
	\end{tabular}
	}
	\caption{\label{table7}Correlation performance of initializing different proportion of negative queue with different negative queue size.}
\end{table}

\begin{table}[tb]
	\centering
	\scalebox{0.9}{
		\begin{tabular}{llllll} 
			\hline
			Corr. & 
			\begin{tabular}[c]{@{}l@{}} 0$\sim$ \\ 512  \end{tabular} & 
			\begin{tabular}[c]{@{}l@{}} 256$\sim$ \\ 768  \end{tabular} & 
			\begin{tabular}[c]{@{}l@{}} 512$\sim$ \\ 1024  \end{tabular} &
			\begin{tabular}[c]{@{}l@{}} Without \\ 256$\sim$768  \end{tabular} & 
			\begin{tabular}[c]{@{}l@{}} All  \end{tabular} \\ 
			\hline
			Avg.    & 76.10   & \textbf{77.02}   & 75.71  & 76.18  & 76.86      \\
			\hline
		\end{tabular}
	}
	\caption{\label{table8}The impact of negative samples at different locations in the queue on the model performance.}
\end{table}

Since the queue is first-in-first-out, to test the hypothesis above, we sliced the negative sample queue and use different parts of the queue to participate in loss calculation. Here, we set the negative queue length to 1024, the initial queue size to 128, and the batch size to 256. Thus, 256 negative samples will be push into the queue for each iteration. We take $0\sim512$, $256\sim768$, $512\sim1024$, a concatenated of slice $0\sim256$ and $768\sim1024$, and all negative sample queues respectively for testing. The experiment results are shown in Table \ref{table8}.

The experiments show that the model performs best when using the middle part of the queue. So we find that the increase in queue length affects the model performance not only because of the increased number of negative samples, but more because it provides historical information within a certain range. 

\subsection{Maximum Traceable Distance Metric}
\begin{figure}[tb]
	\centering
	\includegraphics[width=0.43\textwidth]{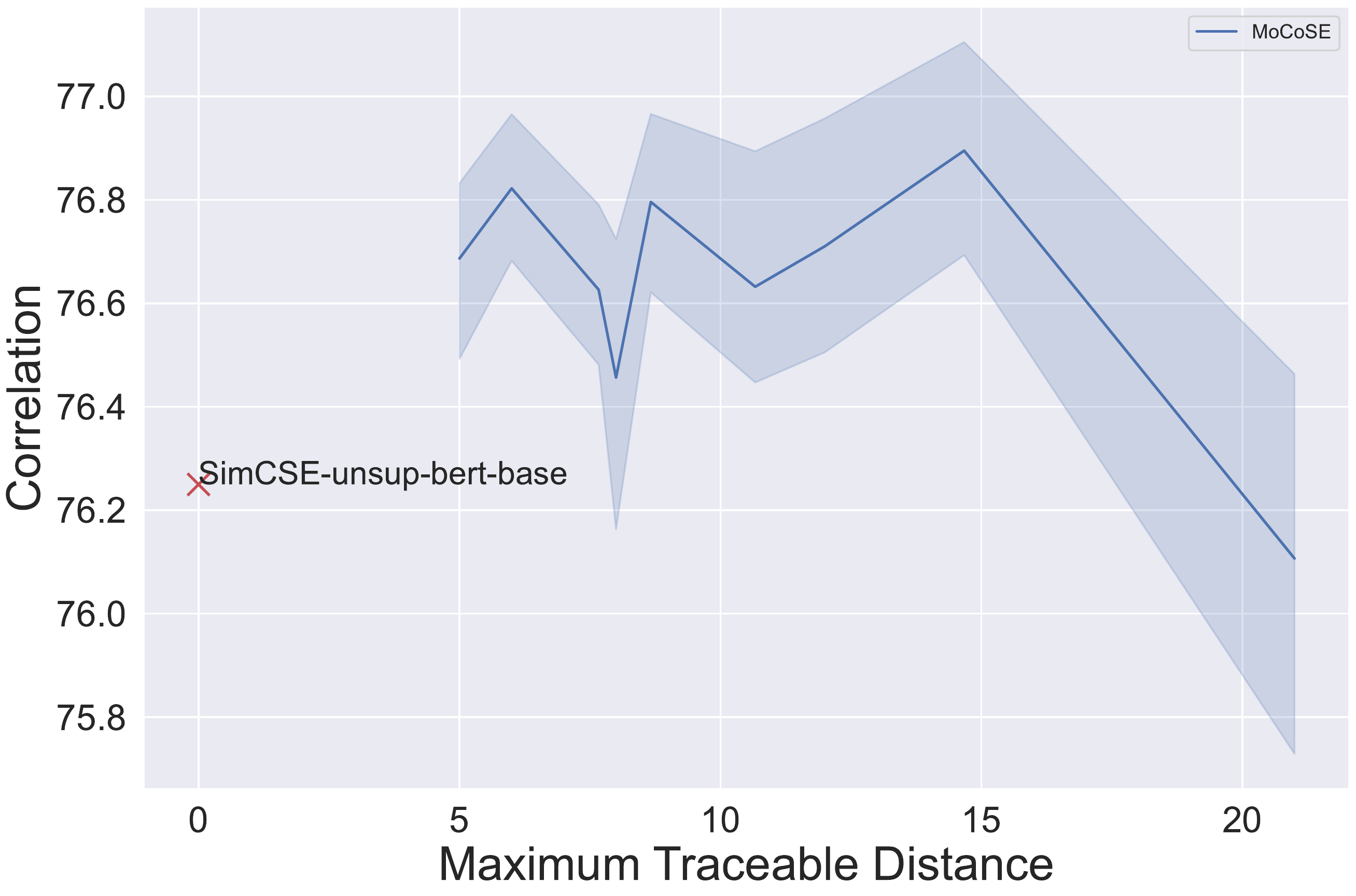}
	\caption{\label{fig2}The relationship between traceable distance and model correlation.}
\end{figure}

\begin{figure}[tb]
	\centering
	\includegraphics[width=0.43\textwidth]{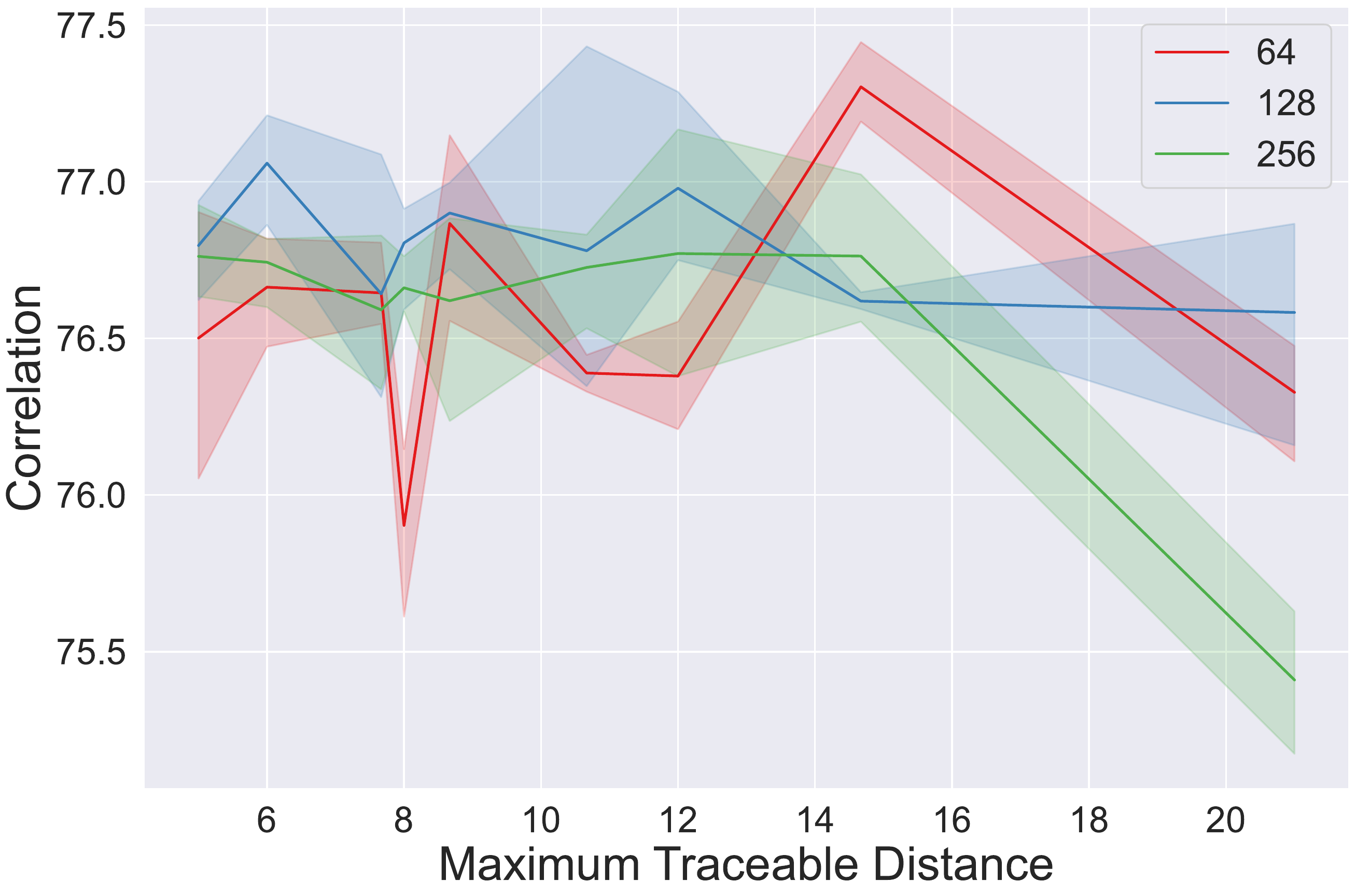}
	\caption{\label{fig3}The batch size does not invalidate the traceable distance. The traceable distance needs to be maintained within a reasonable range even for different batch sizes. This explains why increasing the batch size only does not improve the performance, because increasing the batch size only can cause the distance changes into unsuitable regions.}
\end{figure}

To testify there are historical information in negative sample queue influencing the model performance, we define a Maximum Traceable Distance Metric $d_{trace}$ to help explore the phenomenon.
\begin{equation}
	\begin{aligned}
		d_{trace}=\frac{1}{1-\eta}+\frac{queue\_size}{batch\_size}
		\label{distance}
	\end{aligned}
\end{equation} 
The $\eta$ refers to the decay weight of EMA.
The $d_{trace}$ calculates the update steps between the current online branch and the oldest negative samples in the queue. The first term of the formula represents the traceable distance between target and online branch due to the EMA update mechanism. The second term represents the traceable distance between the negative samples in the queue and the current target branch due to the queue’s first-in-first-out mechanism. The longer traceable distance, the wider the temporal range of the historical information contained in the queue. We obtained different value of traceable distance by jointly adjust the decay weight, queue size, and batch size. As shown in Figure \ref{fig2} and Figure \ref{fig3}, the best result of BERT base is obtained with $d_{trace}$ is set around 14.67. The best result of BERT large shows the similar phenomenon, see Appendix \ref{A.gaplarge} for details. This further demonstrates that in text contrastive learning, the historical information used should be not too old and not too new, and the appropriate traceable distance between branches is also important. Some derivations about eq.\ref{distance} can be found in Appendix \ref{A.proof}.

\begin{figure}[htb]
	\centering
	\includegraphics[width=0.43\textwidth]{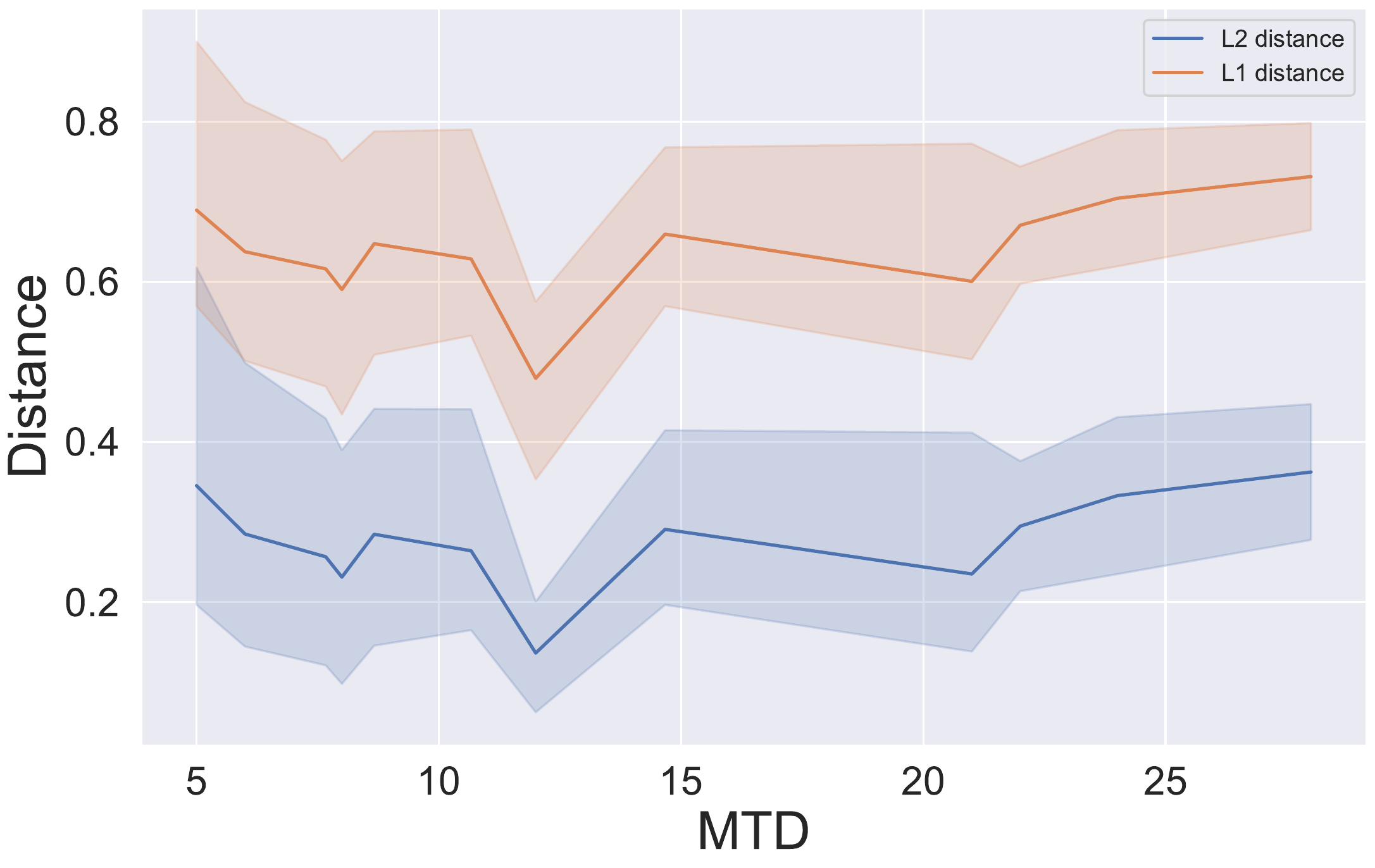}
	\caption{\label{fig_l1l2}L1 and L2 distances of learned embedding's uniformity and alignment with a fixed point changes along with MTD.}
\end{figure}

However, for an image contrast learning model, like MoCo, experimental results suggests that longer queue size increases the performance. We believe that this is due to the phenomenon of unique anisotropy \cite{DBLP:conf/emnlp/0004GXMYS20} of text that causes such differences. The text is influenced by the word frequency producing the phenomenon of anisotropy with uneven distribution, which is different from the near-uniform distribution of pixel points of image data. Such a phenomenon affects the computation of the cosine similarity \cite{wang2020understanding}, and the loss of InfoNCE that we use depends on it, which affects the performance of the model through the accumulation of learning steps. To test such a hypothesis, we use alignment and uniformity to measure the distribution of the representations in space and monitor the corresponding values of alignment and uniformity for different MTDs. As shown in the Figure \ref{fig_l1l2}, it can be found that a proper MTD allows the alignment and uniformity of the model to reflects an optimal combination. The change in MTD is reflected in the performance of uniformity and alignment of the learned text embedding, and the increase and decrease of MTD is a considering result of uniformity and alignment moving away from their optimal combination region.

%% file: sections/conclusion.tex
\section{Conclusion}
In this study, we propose MoCoSE, it applies the MoCo-style contrastive learning model to the empirical study of sentence embedding. We conducted experiments to study every detail of the model to provide some experiences for text contrastive learning. We further delve into the application of the negative sample queue to text contrastive learning and propose a maximum traceable distance metric to explain the relation between the queue size and model performance. 

%% file: sections/appendix.tex
\section{Appendix}

\subsection{\label{A.0} Experiment Settings}
We train our MoCoSE model using a single NVIDIA RTX3090 GPUs. Our training system runs Microsoft Windows 10 with CUDA toolkit 11.1. We use Python 3.8 and PyTorch version v1.8. We build the model with Transformers 4.4.2 \cite{wolf-etal-2020-transformers} and Datasets
1.8.0 \cite{quentin_lhoest_2021_5570305} from Huggingface. We preprocess the training data according to the SimCSE to directly load the stored data in training. We compute the uniformity and alignment metrics of embedding on the STS-B dataset according to the method proposed by Wang \cite{wang2020understanding}. The STS-B dataset is also preprocessed. We use the \textbf{nlpaug} toolkit in our data augmentation experiments. For synonym replace, we use '$ContextualWordEmbsAug$' function with 'roberta-base' as parameter. For typo, we use '$SpellingAug$' and back translation we use '$BackTranslationAug$' with parameter 'facebook/wmt19-en-de' and paraphrase we use '$ContextualWordEmbsForSentenceAug$' with parameter 'xlnet-base-cased'. All the parameter listing here is default value given by official.

\subsection{\label{A.1}Symmetric Two-branch Structure}
\begin{figure}[htb]
	\centering
	\includegraphics[width=0.47\textwidth]{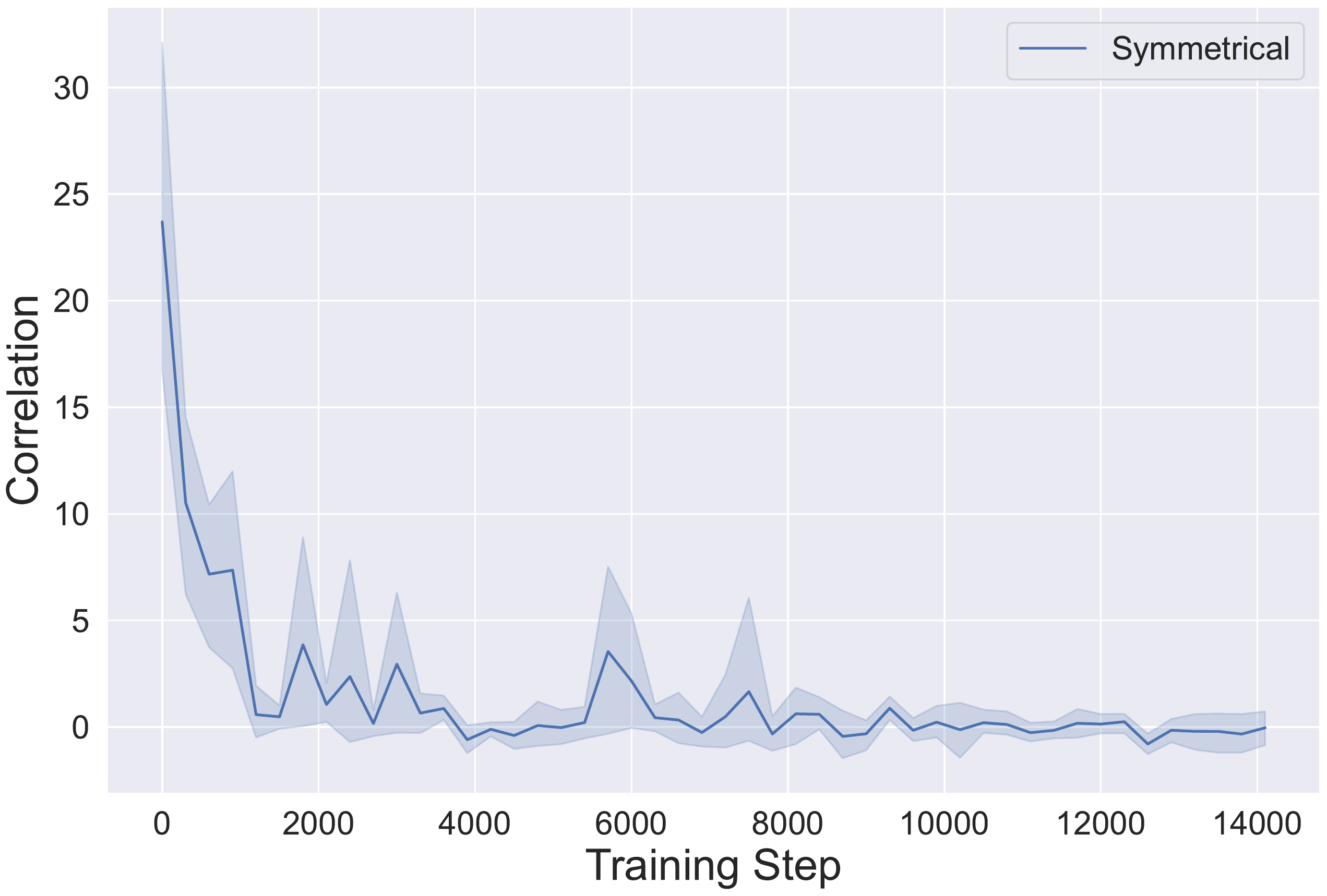}
	\caption{\label{fig:mocose_single}Experiment on a symmetric two-branch structure with EMA decay weight set to 0.}
\end{figure}

We remove the online branch predictor and set the EMA decay weight to 0, i.e., make the structure and weights of the two branches identical. As shown in Figure \ref{fig:mocose_single}, it is clear that the model is collapsing at this point. And we find that the model always works best at the very beginning, i.e., training instead hurts the performance of the model. In addition, as the training proceeds, the correlation coefficient of the model approaches 0, i.e., the prediction results have no correlation with the actual labeling. At this point, it is clear that a collapse of the model is observed. We observed such a result for several runs, so we adopted a strategy of double branching with different structures plus EMA momentum updates in our design. Subsequent experiments demonstrated that this allowed the model to avoid from collapsing.

\begin{figure}[htb]
	\centering
	\includegraphics[width=0.47\textwidth]{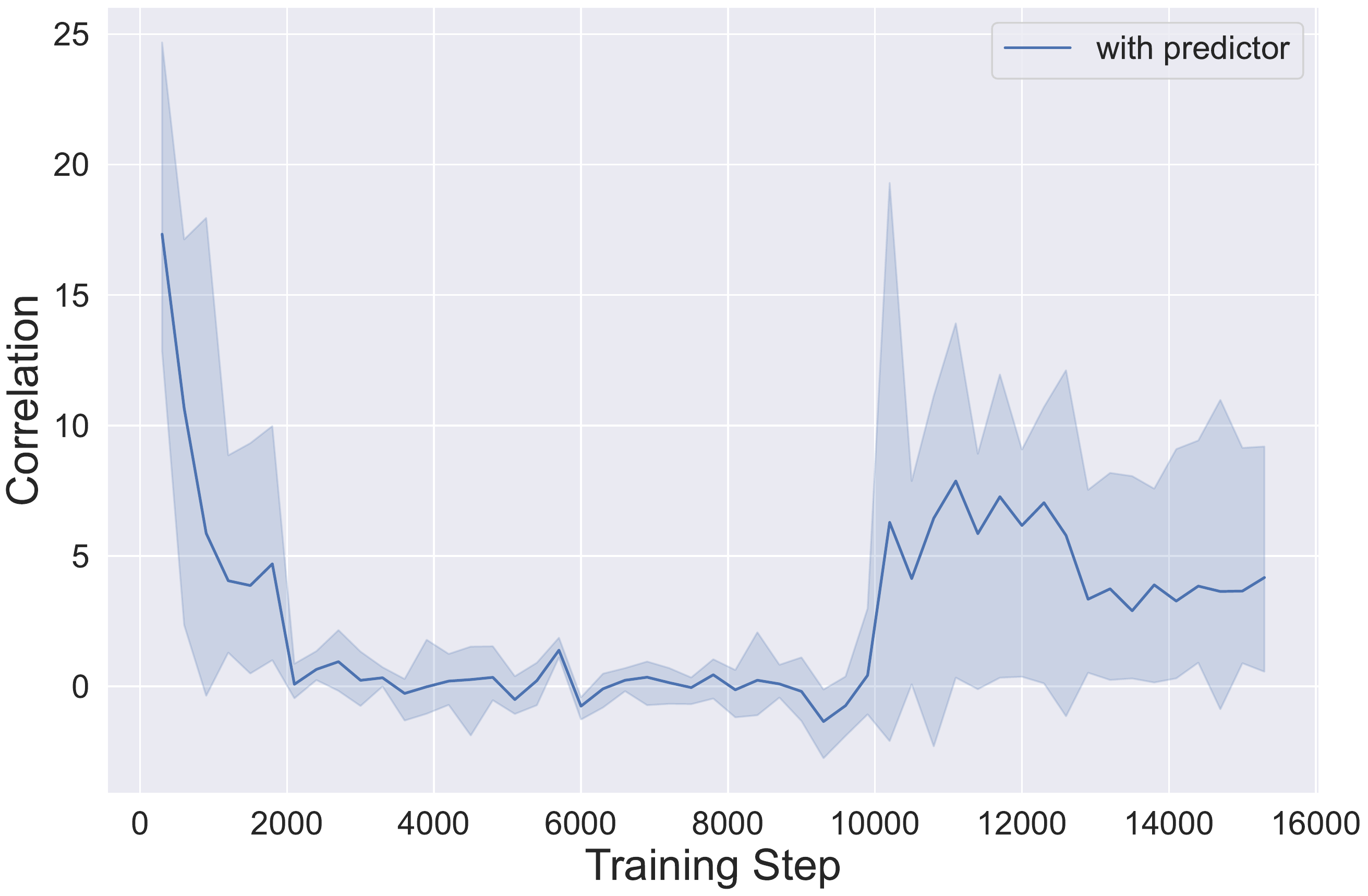}
	\caption{\label{fig:mocose_predictor_collapse}Experiment after adding predictor on the online branch with EMA decay weight set to 0.}
\end{figure}

We add predictor to the online branch and set the EMA decay weight to 0. We find that the model also appears to collapse and has a dramatic oscillation in the late stage of training, as shown in Figure \ref{fig:mocose_predictor_collapse}.

\subsection{\label{A.2}Pseudo-Code for Training MoCoSE}
The PyTorch style pseudo-code for training MoCoSE with the negative sample queue is shown in Algorithm \ref{alg:mocose}.

\begin{algorithm*}[htb]
	\caption{\textbf{Mo}mentum \textbf{Co}ntrastive \textbf{S}entence \textbf{E}mbedding}
	\label{alg:mocose}
	\KwIn{
		
		$\mathcal{D}$ : Training data set  ; \\
		$\mathcal{Q}$ : Negative Sample Queue;\\
		$E_a$ : Embedding with random data augmentation;\\
		$\theta_o , \theta_t$ : weights of online branch and target branch;\\
		Optimizer : Adam optimizer\\
		$K,K_s$: Queue size, Queue size at initialisation;\\
		$\eta$ : ema decay ema and ema scheduling strategy;\\
		$\tau$ Temperature parameters
	}
	\KwOut
	{
		MoCoSE model $\theta_o$\\
	}
	Initializing the queue $\mathcal{Q}$ with size $K_s$; \\
	
	\ForEach{$\mathcal{B} \in \mathcal{D}$}
	{
		\emph{$v_o,v_t \gets E_a\left(\mathcal{B}\right),E_a\left(\mathcal{B}\right)$ \tcp{Using data Augmentation to generate different views}} 
		\emph{$z_o \gets \theta_o\left(v_o\right)$} \tcp{$\left(N,d\right)$, $N$ is batch size, $d$ is dimension of sentence embedding}
		\emph{$z_t \gets \theta_t\left(v_t\right)$} \\
		\emph{$l_{z_o,z_t,\mathcal{Q}} \gets -\log\frac{\exp{\left(z_o \cdot z_t / \tau\right)}}{ \exp{\left(z_o \cdot z_t / \tau \right)} +  \sum_{x \in \mathcal{Q}}\exp{\left(z_o \cdot x / \tau\right)} }$} \tcp{compute contrastive loss using $InFoNCE$}
		\emph{optimizer$\left(l_{z_o,z_t,\mathcal{Q}},\theta_o\right)$} \tcp{Update only the parameters of the online branch according to the loss gradient;}
		\emph{$\theta_t \gets \eta \ast \theta_t + (1-\eta) \ast \theta_o$ } \tcp{Update the parameters of the target branch using $EMA$}
		\emph{enqueue$\left(\mathcal{Q}, v_t\right)$} \tcp{Update the negative sample queue $\mathcal{Q}$}
		\emph{dequeue$\left(\mathcal{Q}\right)$} \\
	}
	return $\theta_o$\\
	
\end{algorithm*}


\subsection{\label{A.3}Distribution of Singular Values}
\begin{figure}[htb]
	\centering
	\includegraphics[width=0.45\textwidth]{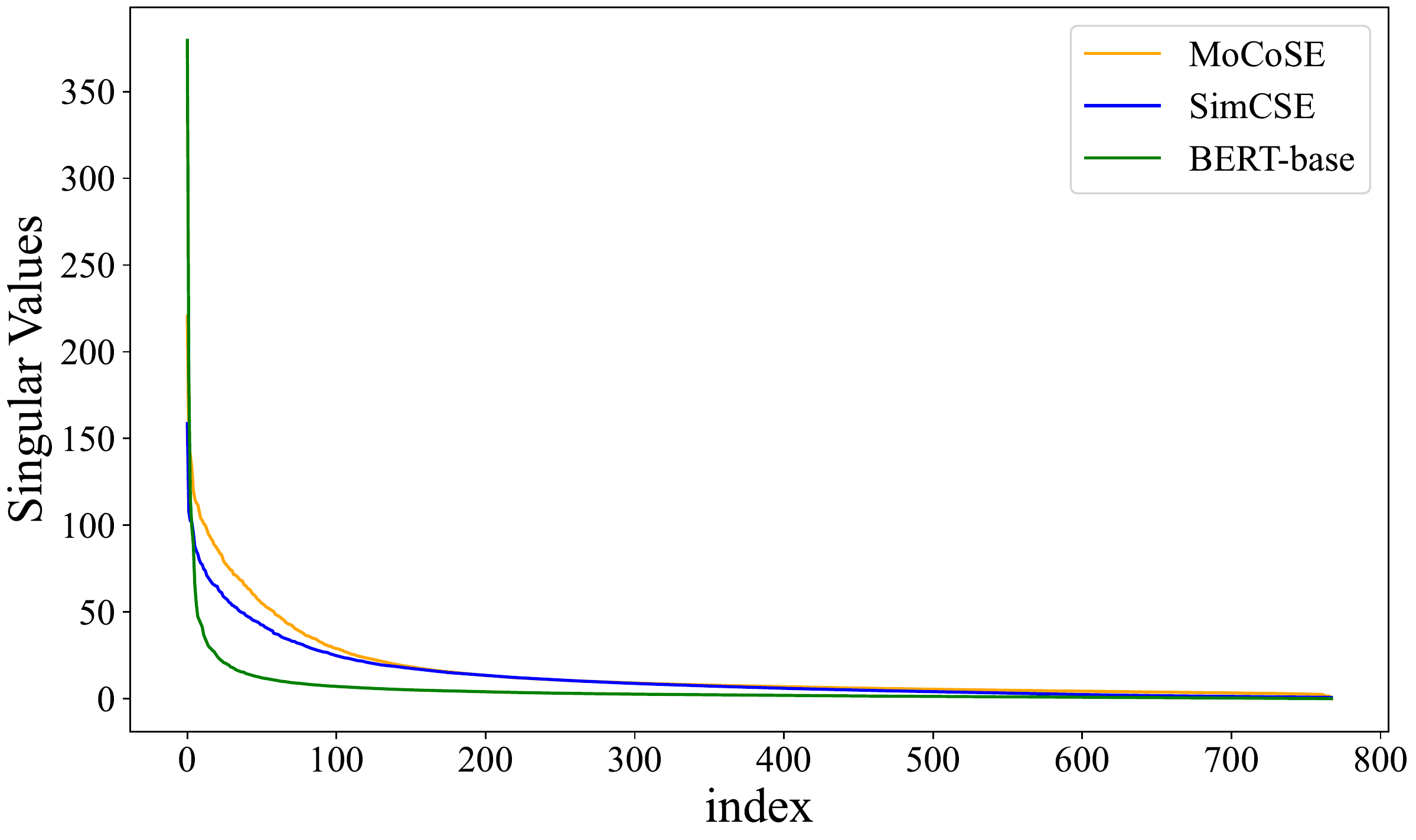}
	\caption{\label{fig:singular_values}Singular value distributions of sentence embedding matrix from sentences in STS-B.}
\end{figure}

Similar to SimCSE, we plot the distribution of singular values of MoCoSE sentence embeddings with SimCSE and BERT for comparison. As illustrated in Figure \ref{fig:singular_values}, our method is able to alleviate the rapid decline of singular values compared to other methods, making the curve smoother, i.e., our model is able to make the sentence embedding more isotropic.

\subsection{\label{A.4}Experiment Details of EMA Hyperparameters }
\begin{figure}[htb]
	\centering
	\includegraphics[width=0.45\textwidth]{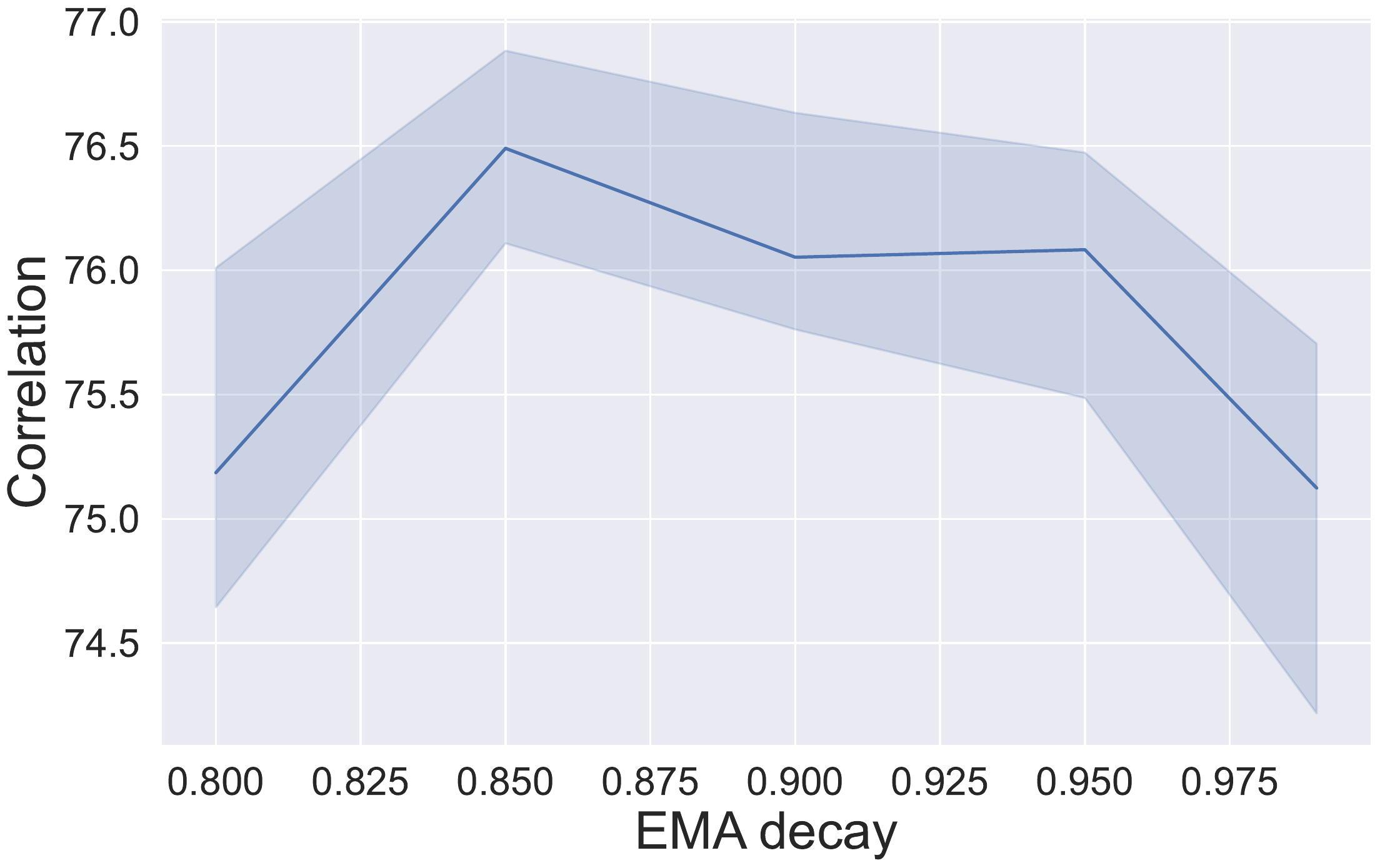}
	\caption{\label{fig:mocose_ema}Effect of EMA decay weight on model performance. }
\end{figure}

The details of the impact caused by the EMA parameter are shown in the Figure \ref{fig:mocose_ema}. We perform this experiment with all parameters held constant except for the EMA decay weight.

\subsection{\label{A.5}Details of Different Data Augmentations}
\begin{figure}[htb]
	\centering
	\includegraphics[width=0.45\textwidth]{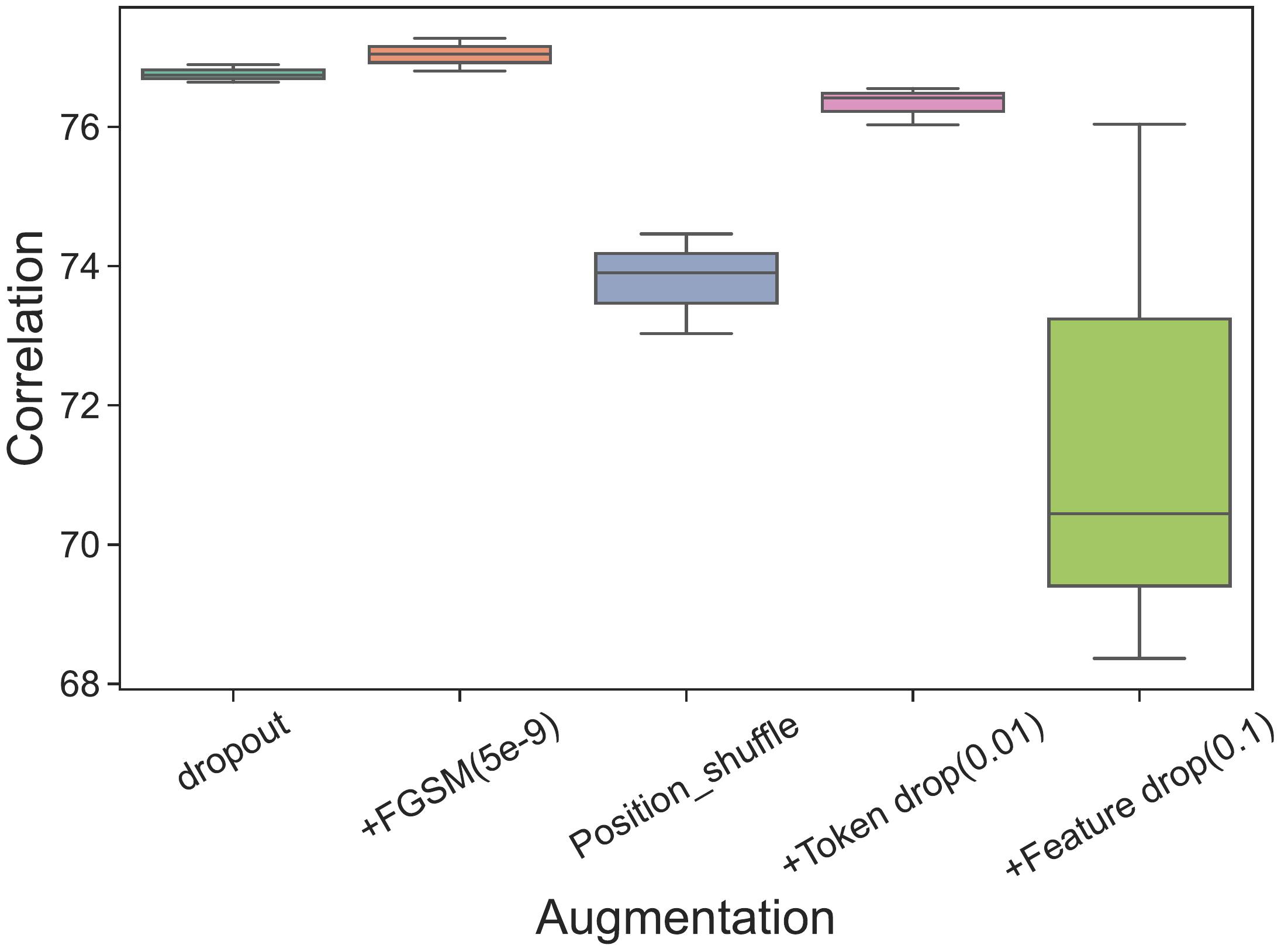}
	\caption{\label{fig:mocose_aug}Impact of four additional data enhancements with dropout combinations on the model. }
\end{figure}

We use only dropout as a baseline for the results of data augmentations. Then, we combine dropout with other data augmentation methods and study their effects on model performance. The results are shown in Figure \ref{fig:mocose_aug}.

\subsection{\label{A.6}Experiment Details of FGSM}
We test the effect of the intensity of FGSM on the model performance. We keep the other hyperparameters fixed, vary the FGSM parameters (1e-9, 5e-9, 1e-8, 5e-8). As seen in Table \ref{table_a2}, the average results of the model are optimal when the FGSM parameter is 5e-9.

\begin{table}[htb]
	\centering
	\begin{tabular}{llllll} 
		\hline
		Epsilon & 1e-9  & \textbf{5e-9}  & 1e-8  & 5e-8  & No     \\ 
		\hline
		Avg.    & 75.61 & \textbf{76.64} & 75.39 & 76.62 & 76.26  \\
		\hline
	\end{tabular}
	\caption{\label{table_a2}Different parameters of FGSM in data augmentation affect the model results.}
\end{table}

\subsection{\label{A.7}Dimension of Sentence Embedding }
\begin{figure}[htb]
	\centering
	\includegraphics[width=0.45\textwidth]{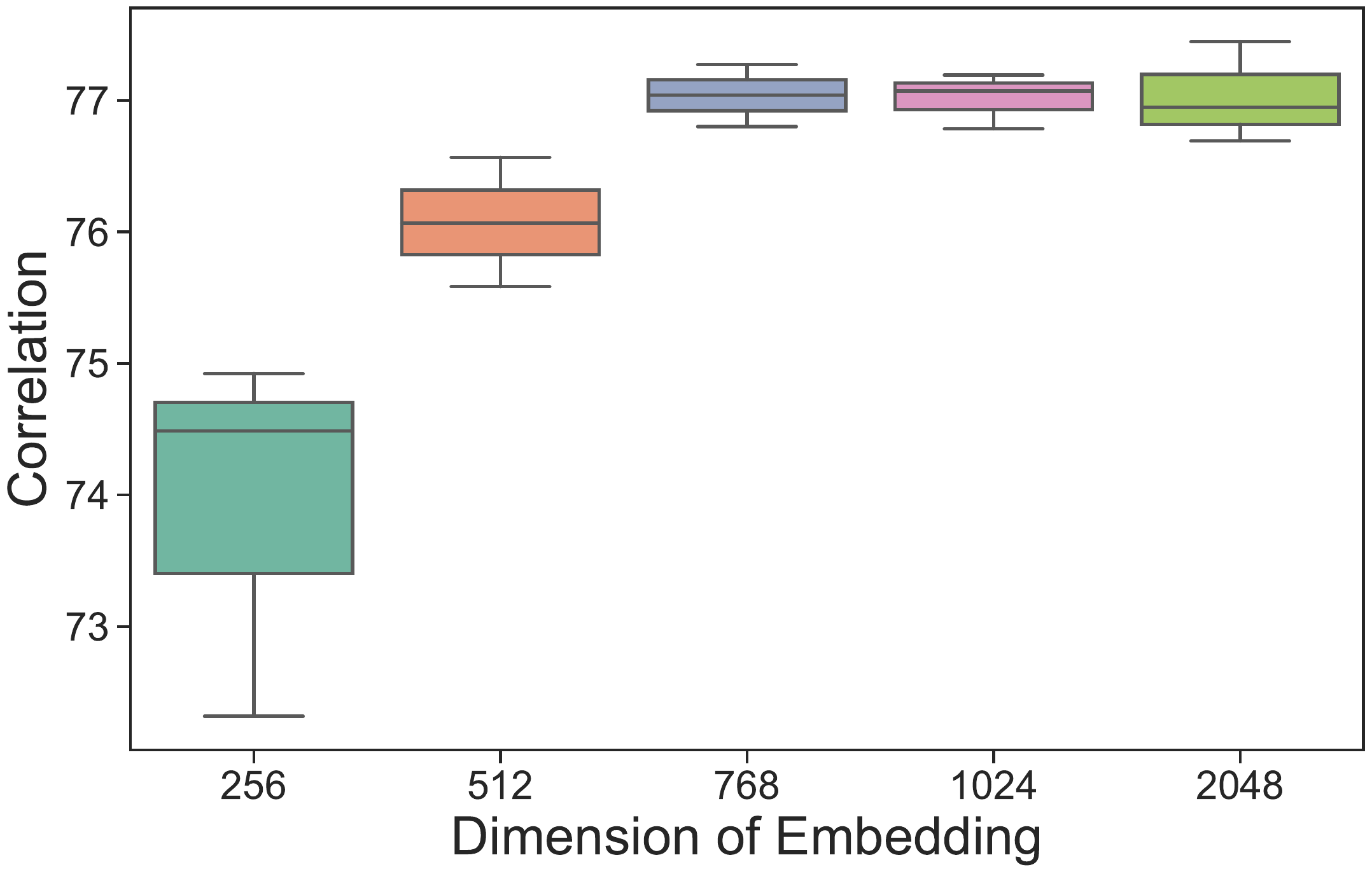}
	\caption{\label{fig:mocose_dim}Impact of dimensions of the sentence embedding.}
\end{figure}

In both BERT-whitening \cite{su2021whitening} and MoCo \cite{he2020momentum}, it is mentioned that the dimension of embedding can have some impact on the performance of the model. Therefore, we also changed the dimension of sentence embedding in MoCoSE and trained the model several times to observe the impact of the embedding dimension. Because of the queue structure of MoCoSE, we need to keep the dimension of negative examples consistent while changing the dimension of sentence embedding. As shown in the Figure \ref{fig:mocose_dim}, when the dimension of Embedding is low, this causes considerable damage to the performance of the model; while when the dimension rises to certain range, the performance of the model stays steady.

\subsection{\label{A.8}Details of Random Initial Queue Size}
\begin{figure}[htb]
	\centering
	\includegraphics[width=0.45\textwidth]{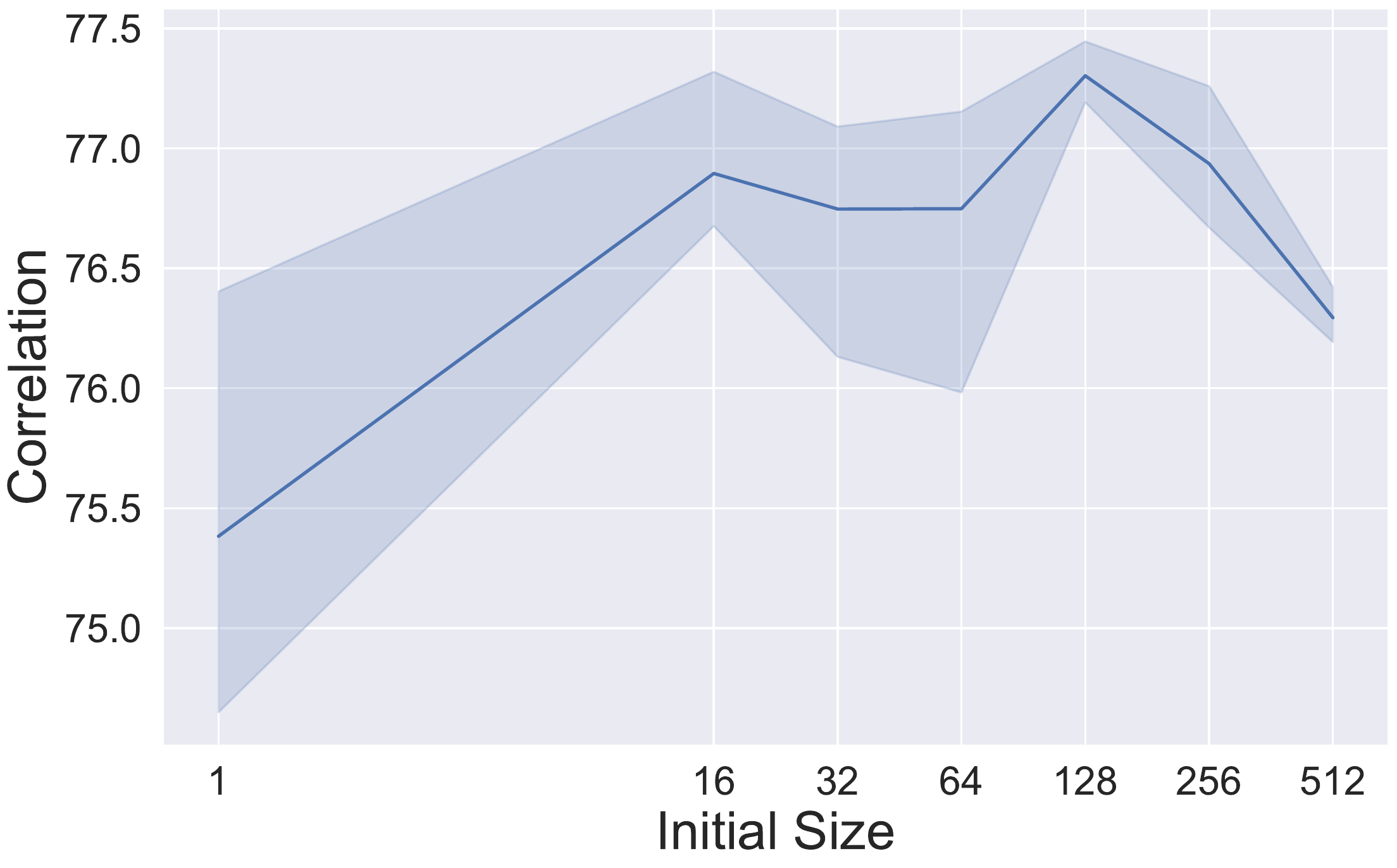}
	\caption{\label{fig:mocose_init}The effect of the initial queue size on the model results when the queue length is 512 and the batch size is 64. }
\end{figure}

We test the influence of random initialization size of the negative queue on the model performance when queue length and batch size are fixed. As seen in Figure \ref{fig:mocose_init}, random initialization does have some impact on the model performance.

\subsection{\label{A.9}Queue Size and Initial Size}
\begin{figure*}[htb]
	\centering
	\includegraphics[width=0.85\textwidth]{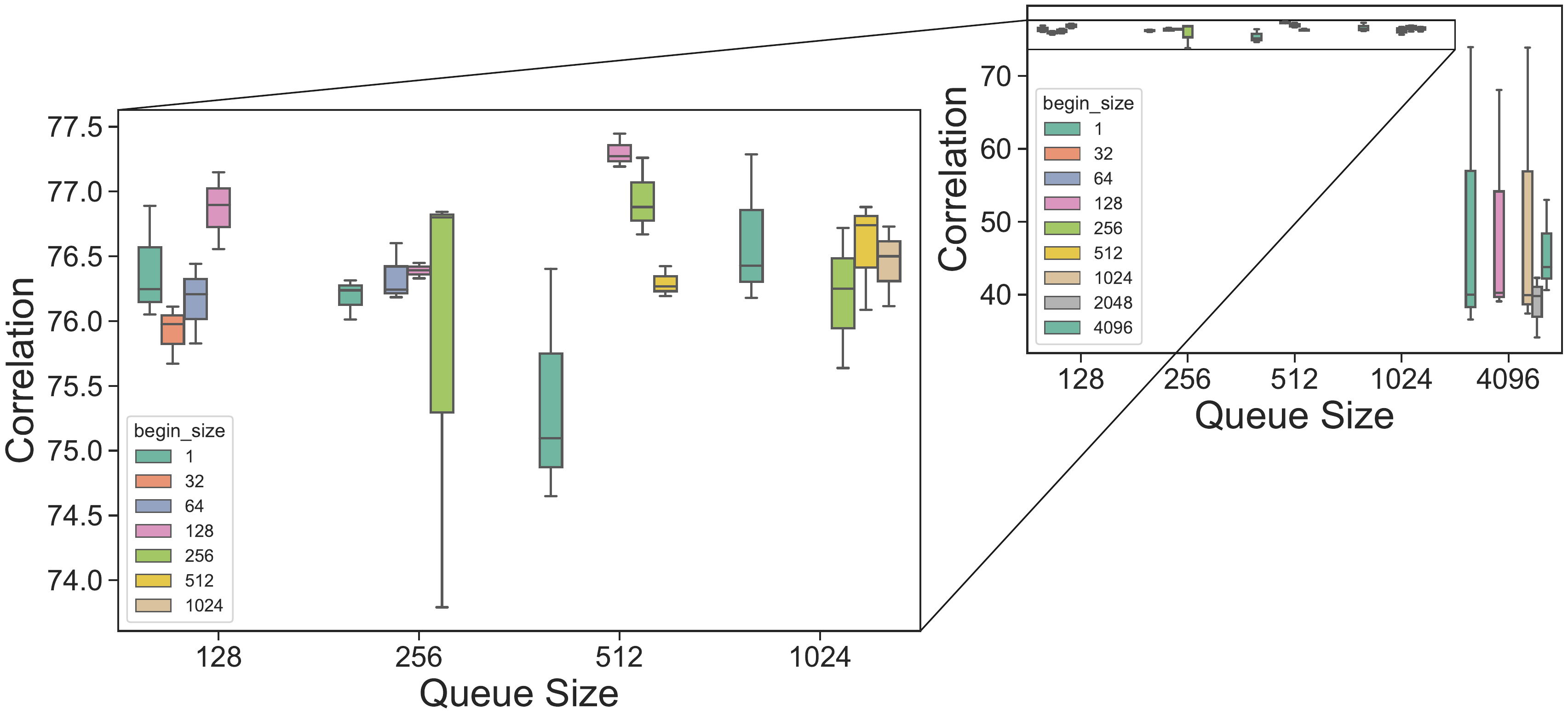}
	\caption{\label{fig:mocose_queuet}The impact of different initial negative sample queue sizes for different initial  sizes on model performance. (left):Zoomed view. (right):Overview with different negative queue size. Results of different initial size under same queue size.}
\end{figure*}

We explored the effect of different combinations of initial queue sizes and queue length on the model performance. The detailed experiment results are shown in Figure \ref{fig:mocose_queuet}. It can be found that model performance rely deeply on initialization queue size. Yet, too large queue size will make the model extremely unstable. This is quite different from the observation of negative sample queue in image contrastive learning.

\subsection{\label{A.gaplarge}Maximum Traceable Distance in BERT-large}
\begin{figure}[htb]
	\centering
	\includegraphics[width=0.45\textwidth]{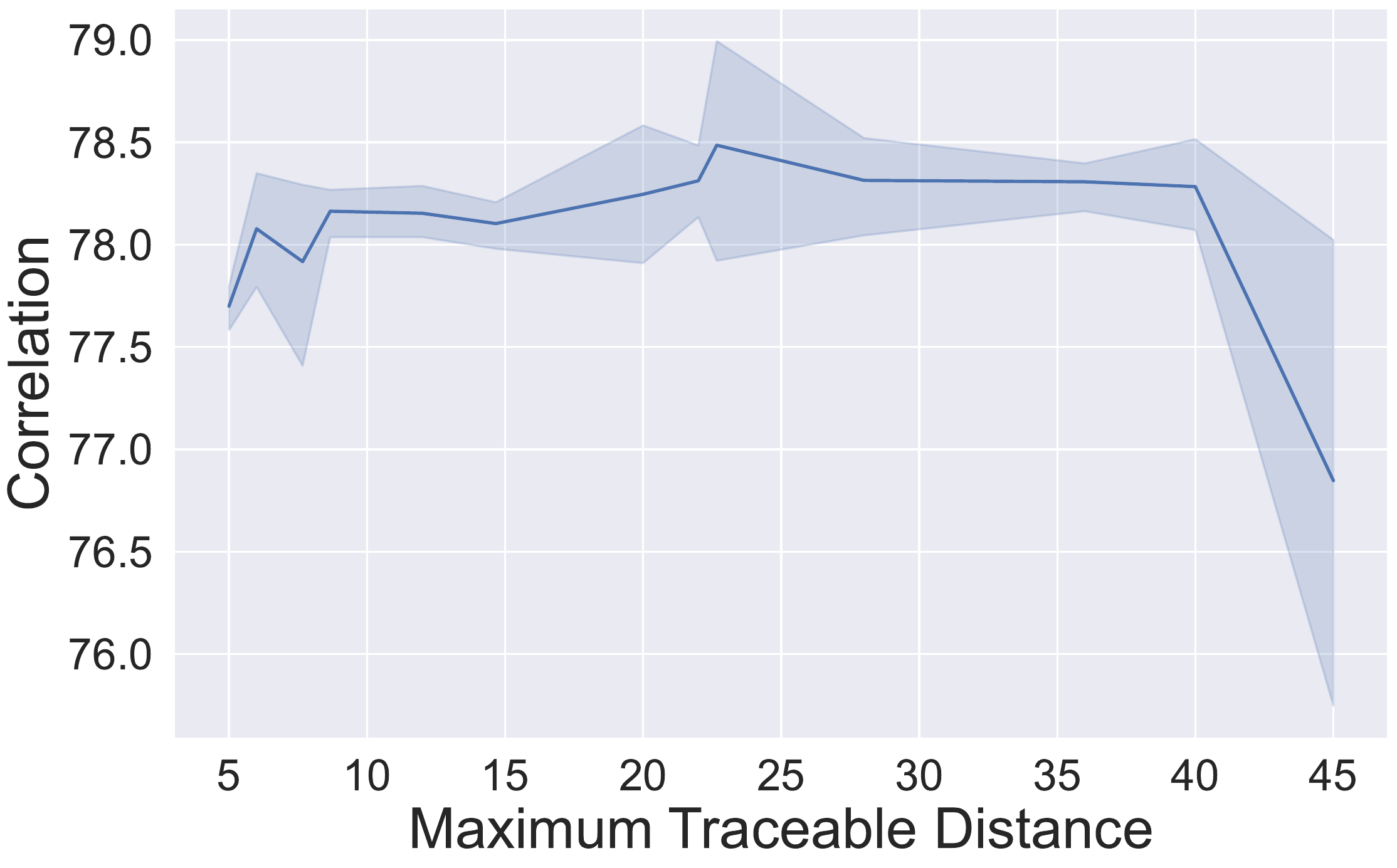}
	\caption{\label{fig:mocose_gap_all_large}The relationship between MTD and correlation of MoCoSE-BERT-large. It can be seen that even at large model, peaks occur within a certain MTD range.}
\end{figure}

We also train mocose with different batch size and queue size on BERT-large. As shown in Figure \ref{fig:mocose_gap_all_large}, we observe the best model performance in MoCoSE-BERT-large within the appropriate Maximum Traceable Distance range (around 22). Once again, this suggests that even on BERT-large, the longer queue sizes do not improve the model performance indefinitely. Which also implies that the history information contained in the negative sample queue needs to be kept within a certain range on BERT-large as well.


\subsection{\label{A.proof}Proof of Maximum Traceable Distance}
Here, we prove the first term of the formula for Maximum Traceable Distance. Due to the EMA update mechanism, the weight of target branch is a weighted sum of the online weight in update history. The first term of Maximum Traceable Distance calculate the weighted sum of the historical update steps given a certain EMA decay weight $\eta$.
From the principle of EMA mechanism, we can get the following equation.

\begin{equation}
	\label{eq:sum_sn_1}
	\begin{aligned}
		\mathcal{S}_n = \sum_{i=0}^{k}\left(1-\eta\right) \cdot \eta^{i} \cdot \left(i+1\right)
	\end{aligned}
\end{equation}
$\mathcal{S}_n$ represents the update steps between online and target branch due to the EMA mechanism. Since EMA represents the weighted sum, we need to ask for $\mathcal{S}_n$ to get the weighted sum. \\

We can calculate $\mathcal{S}_n$ as:
\begin{equation}
	\label{eq:sum_sn_7}
	\begin{aligned}
		\mathcal{S}_n &= \left(-1\right) * \eta^{k+1} * \left(k+1\right) - \frac{\left(1-\eta^{k+1}\right)}{\left(\eta-1\right)}
	\end{aligned}
\end{equation}
As $k$ tends to infinity, the limit for $\mathcal{S}_n$ can be calculated as following:
\begin{equation}
	\label{eq:sum_sn_8}
	\begin{aligned}
		& \lim_{k\to\infty} \mathcal{S}_n = \\ 
		& \lim_{k\to\infty} \left[\left(-1\right) * \eta^{k+1} * \left(k+1\right) - \frac{\left(1-\eta^{k+1}\right)}{\left(\eta-1\right)}\right]
	\end{aligned}
\end{equation}
It is obvious to see that the limit of the equation \ref{eq:sum_sn_8} consists of two parts, so we calculate the limit of these two parts first.
\begin{equation}
	\label{eq:sum_sn_9}
	\begin{aligned}
		\lim_{k\to\infty} \left(-1\right) * \eta^{k+1} * \left(k+1\right) \overset{\eta < 1}{=} 0
	\end{aligned}
\end{equation}
The limit of the first part can be calculated as $0$. Next, we calculate the limit of the second part.
\begin{equation}
	\label{eq:sum_sn_10}
	\begin{aligned}
		\lim_{k\to\infty} \frac{\left(1-\eta^{k+1}\right)}{\left(\eta-1\right)} \overset{\eta < 1}{=} \frac{1}{1-\eta}
	\end{aligned}
\end{equation}
We calculate the limit of the second part as $\frac{1}{1-\eta}$. Since the limits of both parts exist, we can obtain the limit of $\mathcal{S}_n$ by the law of limit operations.
\begin{equation}
	\label{eq:sum_sn_11}
	\begin{aligned}
		& \lim_{k\to\infty} \mathcal{S}_n \\ 
		& = \lim_{k\to\infty} \left[\left(-1\right) * \eta^{k+1} * \left(k+1\right) - \frac{\left(1-\eta^{k+1}\right)}{\left(\eta-1\right)}\right] \\
		& = \lim_{k\to\infty} \left(-1\right) * \eta^{k+1} * \left(k+1\right) - \lim_{k\to\infty}\frac{\left(1-\eta^{k+1}\right)}{\left(\eta-1\right)} \\
		& = \frac{1}{1-\eta}
	\end{aligned}
\end{equation}

%